\documentclass[runningheads]{llncs}
\usepackage{pifont}% http://ctan.org/pkg/pifont
\usepackage[linesnumbered,ruled,resetcount]{algorithm2e}
\usepackage{graphicx}
\usepackage{rotating,amsmath}
\usepackage{multirow}
\usepackage{todonotes}
\usepackage{color}

\begin{document}
\title{Impact of meta-roles on the evolution of organisational institutions }
\titlerunning{Impact of meta-roles on the evolution of institutions}

\author{Amir Hosein Afshar Sedigh\inst{1}\and Martin K. Purvis\inst{1} \and Bastin Tony Roy Savarimuthu\inst{1}\and Maryam A. Purvis\inst{1}\and Christopher K. Frantz\inst{2}}

\authorrunning{A. H. Afshar Sedigh et al.}
%\institute{}
\institute{Department of Information Science, University of Otago, Dunedin, New Zealand \and Department of Computer Science, Norwegian University of Science and Technology, Ametyst-bygget, A205, Gj{\o}vik, Norway \email{amir.afshar@postgrad.otago.ac.nz}\\ \email{\{martin.purvis,tony.savarimuthu,maryam.purvis\}@otago.ac.nz}\\
%\url{http://www.springer.com/gp/computer-science/lncs} 
\email{christopher.frantz@ntnu.no}}
%
% Comments
\maketitle              % typeset the header of the contribution
\begin{abstract}
This paper investigates the impact of changes in agents' beliefs coupled with dynamics in agents' meta-roles on the evolution of institutions. The study embeds agents' meta-roles in the BDI architecture. In this context, the study scrutinises the impact of cognitive dissonance in agents due to unfairness of institutions. To showcase our model, two historical long-distance trading societies, namely Armenian merchants of New-Julfa and the English East India Company are simulated. Results show how change in roles of agents coupled with specific institutional characteristics leads to changes of the rules in the system.
\keywords{Institutions  \and BDI \and Agent-based simulation \and Meta-roles \and Cognitive dissonance}
\end{abstract}
\section{Introduction} 
Employing evolutionary methods to study economic change has attracted several scholars. For instance, Nelson and Winter proposed the idea that ``organisational routines'' are pivotal in the evolution of business firms (i.e.~their role is similar to the role of the genes in biological evolution) \cite{OrgEvol}. Also, they suggested that ``[metaphorically] [r]outines are the skills of an organisation.'' However, different scholars suggested various definitions of routines. For instance, Feldman and Pentland called ``a repetitive, recognizable pattern of interdependent actions, involving multiple actors'', a routine \cite{Feldman2003}. %Overall, routines have some attributes in common; they are recurrent, unlike skills, they are collective phenomena, they have a processual nature, and they are context-dependent \cite{Becker2004}.
Routines have similarities with institutions (e.g.~`the rules of the game' \cite{North1991}), in terms of their collective attributes \cite{OrganRoutInstitu} (i.e.~they have rule-like conditions \cite{Hodgson2003}). However, whether routines are subconsciously followed (they are simple rules) or they are open to amendments and changes (they are ambiguous rules) is subject of controversy \cite{Becker2004}. Hodgson \cite{Hodgson2003} criticised Nelson and Winter's \cite{OrgEvol} method and pointed to some shortcomings such as not considering `birth' and `death' in their method.

Also, in computer science, role/meta-role based frameworks were developed to facilitate modelling. For instance, Riehle and Gross \cite{Riehle1998} developed a role modelling approach `to describe the complexity of object collaborations.' Also, MetaRole-Based Modelling Language (RBML) was expressed in the Unified Modeling Language (UML) to describe patterns' attributes \cite{RBML}. The CKSW (\textit{Commander}--textit{Knowledge--Skills--Worker}) framework was proposed for meta-role modelling in agent-based simulation \cite{Purvis2014}. The idea of integrating roles and institutions is already studied in the context of multi-agent systems. For instance, nested ADICO refined Ostrom's grammar of institutions \cite{Crawford1995} by differentiating between roles (e.g.~enforcer and monitoring agent) \cite{frantz2013nadico}.

 The BDI (beliefs-desires-intention) model is a cognitive agent architecture \cite{Bratman1988} with some extensions, including the BOID \cite{BOID}, EBDI \cite{Pereira2008} and the BRIDGE \cite{Dignum2009} models. This architecture was employed to model agents' cooperation in institutionalised multi-agent systems \cite{BDIInst,Balke2012}. 
 
 In light of earlier studies, this paper integrates agents' meta-roles \cite{Purvis2014} in the BDI architecture and also employs the theory of planned behaviour TPB \cite{fishbein2011predicting} to model different facets of beliefs. The integrated model is used to investigate how dynamics in agents’ meta-roles may lead to the evolution of organisational institutions. Meta-roles in this work are modelled using the CKSW framework that helps modellers to decompose agents in a society based on the characteristics of their roles \cite{Purvis2014}. The coupling of the CKSW framework within a BDI architecture is investigated in the context of rule-making and -following (how rules are established, interpreted, and followed). 

\section{An overview of the extended BDI architecture}
\label{cogarc}

\begin{figure}[hbt!]
		\centering
			\includegraphics[trim=.5cm 13.55cm 2.1cm 3.3cm,clip, width=1\textwidth]{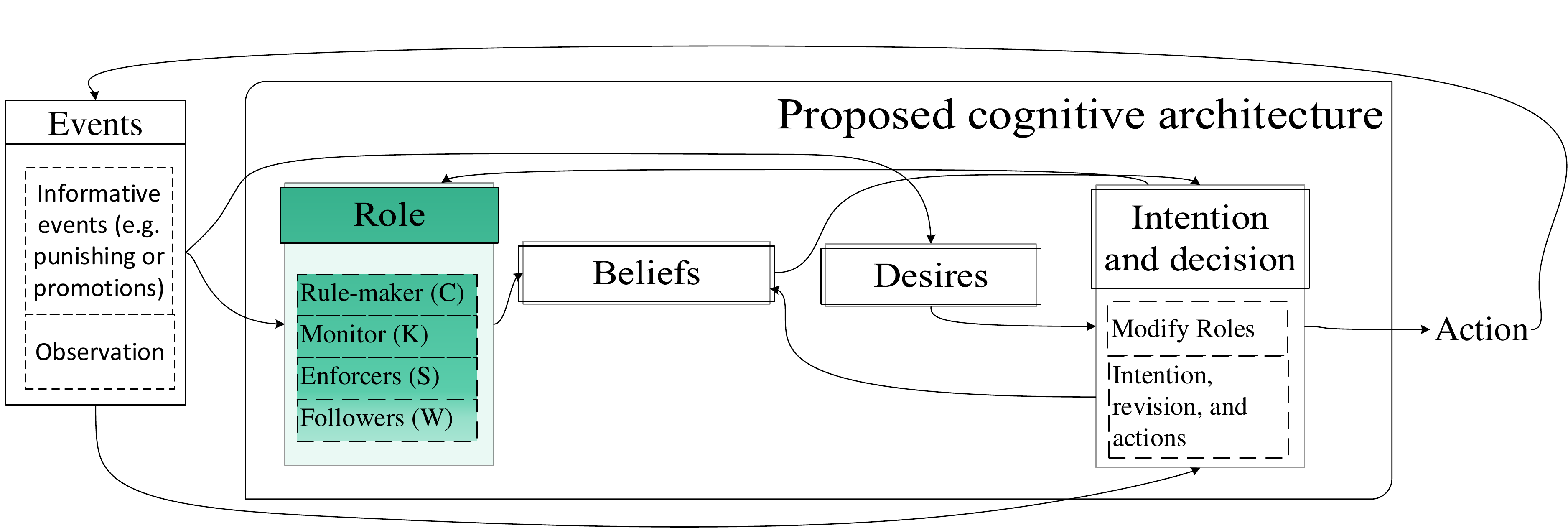}
		\caption{Proposed cognitive architecture for this model.}
		\label{architec}
	\end{figure}

 This extended BDI cognitive architecture is shown in Figure \ref{architec}. It can be observed that there are two separate blocks, a left block called `\textit{Events}' and a right block called `{Cognitive architecture}'. The {Events} block represents the events an agent perceives from the environment (e.g.~information collected from peers). The \textit{Cognitive architecture} block represents an agent's cognitive decision-making components. Note that when an action is performed by an agent, it will be an input event for those agents interested in that event in the next iteration. 
 A brief description of the four high-level components is provided below. It should be noted that the main focus of this paper is on the addition of \emph{Role} component to the BDI architecture (highlighted in green in Figure \ref{architec}).
 
 	\begin{itemize}%\cmt{Regarding Figure: Why are there empty grey boxes under beliefs and desires. Put them on the same level and adjust the arrows; look cleaner. I would make it clear that this is a high-level view, e.g., by indicating it in the caption. It is very broad and would invite for a lot of questions, e.g., how it is different from existing approaches. Is that something worth discussing?}
				\itemsep0em 
 \item \textbf{Roles:} An agent has a set of roles in society regarding established institutions (e.g.~agents make those institutions or they monitor their implementations). An agent's role impacts its beliefs, based on individual and social experiences (e.g.~it personally may find the rule unfair). We discuss this module in more detail in Section \ref{rolesdynmeta}.%Note that roles include both internalised and associated roles with respect to its position.
 \item \textbf{Beliefs:} To model beliefs, we are inspired by the idea of different belief components of TPB \cite{fishbein2011predicting}. %namely internal beliefs about rules, perceived norms in the system, and its belief about the purpose of the rule (rule-understanding). %Furthermore, an agent has other beliefs (e.g.~its cognition of parameters). What follows discusses the three components of an agent's beliefs, as discussed in Section \ref{TPB:subsec}. These three components comprise: 
 %	\begin{itemize}
 %				\itemsep0em 
 %	\item \textbf{Internal beliefs:} This component indicates an agent's belief about what the rule must be. Breaking this imposes mental costs on an agent, whether or not others observe the action (it is inspired by \textit{behavioural beliefs} discussed in Section \ref{TPB:subsec}). 
 %	\item \textbf{The perceived norms:}
 This component indicates an agent's perception about the rule and the support the rule has. In other words, an agent has its own internal belief about the rule, and also the perception about the social support for that rule (e.g.~rebuking the rule), and an estimation of what an organisation meant by the rule (e.g.~consequences of minor violation).
  %perception of 
  %societal support for the rule --- for instance, an agent's perception of possible sanctions for not following organisational rules. Breaking these perceived norms imposes costs on an agent when other agents identify the violation (it is inspired from \textit{normative beliefs} discussed in Section \ref{TPB:subsec}).
% 	\item \textbf{Rule-understanding:} This component represents the rule as an agent understands it. This may differ from the real intention of the rule-maker. This is enforced by agents who have the duty of monitoring, reporting, and punishing the violators. This component has the most rigid punishments, such as dismissal, repaying the costs, and jailing. For executing this component, the system needs some official reports about the agents' behaviour (it is inspired from \textit{control beliefs} discussed in Section \ref{TPB:subsec}).
% \end{itemize}
 \item \textbf{Desires:} Agents have different desires, such as an agent's goals and ideal preferences. 
 \item \textbf{Intentions and decision:} An agent's intentions for an action and its decision about the final action is formed in this module. The decision results in an action which can be a modification of beliefs and roles or only performing a task. %We expand this module in Section \ref{decisionmodule}.
	\end{itemize}
	
\section{Meta-roles and role dynamics}
\label{rolesdynmeta}
To model agents' roles and their interactions we use CKSW meta-roles \cite{Purvis2014}. Note that CKSW is a generic model and since this paper concerns the rule-making and rule-following context, we reinterpret those roles in this context as follows:

\begin{itemize}
	\itemsep0em 
	\item \textit{Commander} (\textbf{C}): This role is empowered with ultimate authority \cite{Purvis2014}. In this context, they are the agents who are permitted to \textit{make or revise rules}.
	\item \textit{Knowledge} (\textbf{K}): This role concerns the \emph{know-what} aspect of a society \cite{Purvis2014}. In this context, these are agents who \textit{monitor and report} the \textit{suspicious activities} of others.
	\item \textit{Skills} (\textbf{S}): This role concerns agents who are known for their skills in society (\emph{know-how}). Unlike knowledge, skills are difficult to communicate and much more so to apply \cite{Purvis2014}. In the rules context, those agents that have the skills to interpret the rules judge reported agents' activities.
	\item \textit{Worker} (\textbf{W}): These agents perform basic jobs that do not require specialist skills \cite{Purvis2014}. In this context, they are agents who do not formally collaborate in monitoring, establishing, or interpreting the rules (i.e.~the rest of agents). 
\end{itemize}

%\begin{figure}%[hbt!]
%	\centering
%	%	trim=left bottom right top
%	\includegraphics[trim=0cm 18cm 5.4cm .5cm,clip, width=.75\textwidth]{meta-roles}
%	\caption{Agents' meta-roles and their relations adapted for rule-following context.}
%	\label{metaroles}
%\end{figure}

%Figure \ref{metaroles} depicts how the aforementioned meta-roles interact in an organisation. Commander agents (e.g.~a board of directors) decide about the rules regarding issues such as trades. They declare those rules to all other agents. An agent's behaviour is monitored by Knowledge agents (e.g.~managers) to inform Skill agents (e.g.~judges) about suspicious actions. Furthermore, based on agents' behaviour, Knowledge agents may suggest some revisions of rules to the Commander agents. Skill agents who also need some elements of Knowledge (i.e.~they need some knowledge besides skills and are labelled $ \mathrm{S^K} $), interpret the rules and available evidence to punish the violators. Commander agents may express objections to some of those interpretations with respect to their powers.

We also consider two categories of roles, \emph{formal roles} and \emph{informal roles}:
\begin{itemize}
	\itemsep0em
	\item Formal roles: these roles are defined based on the agent's position in an organisation (one of CKSW meta-roles).
	\item Informal (internalised) roles: these roles are unofficially and self-assigned (e.g.~based on values) by agents such as monitoring, and reporting suspicious behaviours of other agents to managers. These are the role(s) that an agent may perform in addition to its formal role (one or more out of the CKSW meta-roles).
\end{itemize}

\begin{figure}[hbt!]
	\centering
	%	trim=left bottom right top
	\includegraphics[trim=.5cm 17.5cm 4.4cm .5cm,clip, width=.8\textwidth]{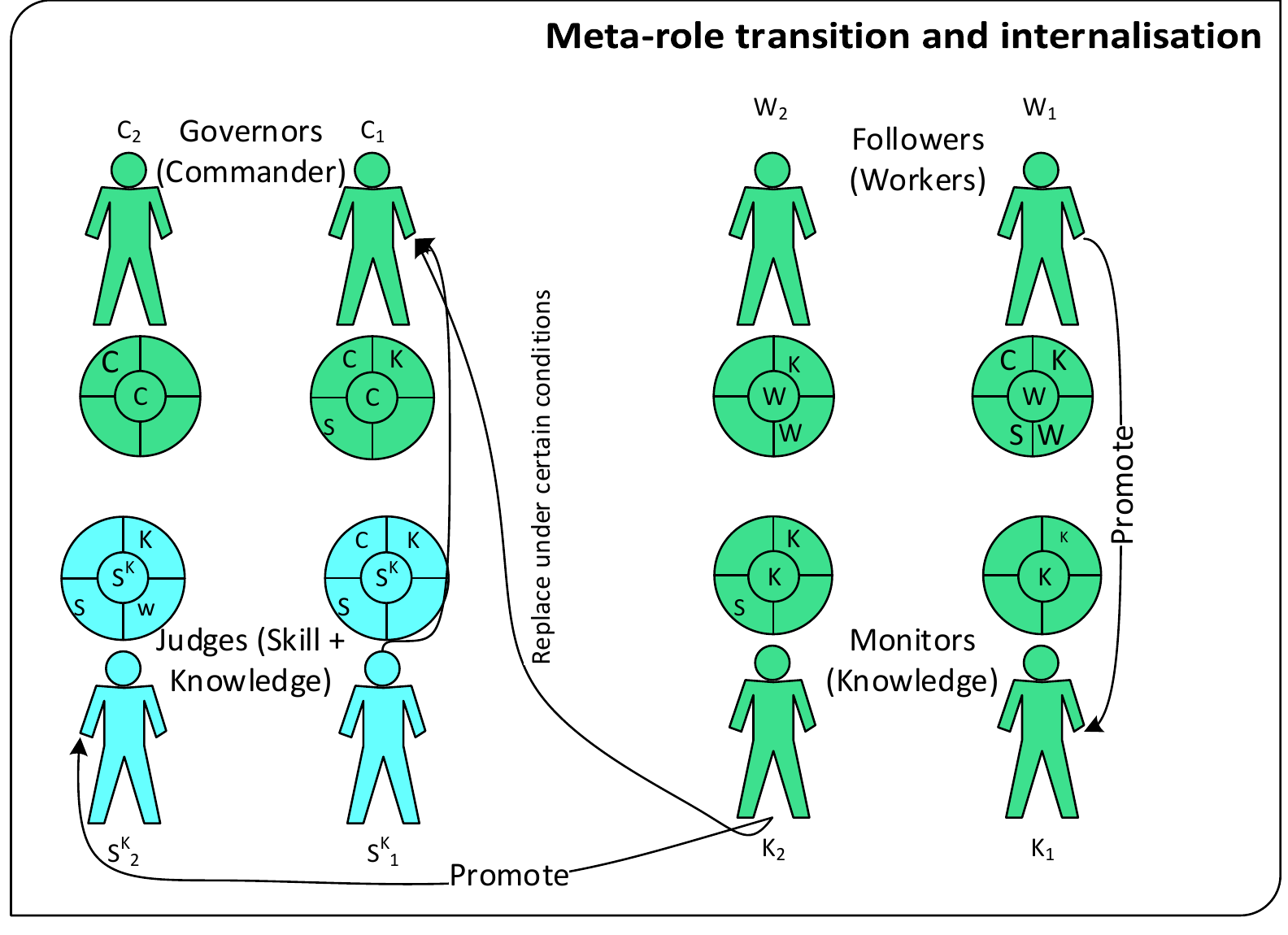}
	\caption[Transition of formal meta-roles in an organisation]{Transition of formal meta-roles and internalising informal meta-roles --- the circles indicate roles that an agent really performs (including internalised roles), and the bigger fonts indicate more involvement in such a role. The arrows indicate the possible transitions in a society.}
	\label{metaroles2}
\end{figure}%\cmt{It should be clarified that this figure is an example. The arrows are generic, but the specific persons are exemplified. That should be clear from the description, else ppl. will argue about the specific. Try to center the type annotations (e.g., C in C2 commander) for better legibility. Also, consider how it looks on black and white. I would suggest white and gray boxes. Easier to parse when printed.}

Figure \ref{metaroles2} depicts how an agent's meta-role may evolve\footnote{Note that most times agents are downgraded for economic issues or bad performance of agents, and this downgrading can be considered as an extension of this model.}. In this example, a worker (say clerk) of an organisation may be promoted to a higher rank after demonstrating competence for such a promotion (say to a manager, a knowledge-based role). If the manager has relevant education, skills and experience, it can be promoted to an even higher position. In these positions, the manager may be responsible for interpreting the situation and deciding about who to fire or hire (i.e.~promotion from Knowledge role to the role of a judge (skill) --- represented as $ \mathrm{S^K} $ in the figure). Under certain conditions an agent (Knowledge or Skill) can be promoted to  director role (i.e.~to the Commander). Note that judges (highlighted in blue) might not be explicitly present on an organisational level for various reasons (e.g.~sometimes legal cases go to international courts).

%The informal roles are voluntarily added to an agent's formal roles (\textit{role internalisation}). 

In Figure \ref{metaroles2}, the initials for formal roles are indicated on top or bottom of each agent (e.g. $K_1$) and the list of all roles for an agent (i.e.~informal and formal roles) are shown in  circles placed near the agents. The font sizes of initials inside these circles indicate the involvement level in such a role, with larger fonts involving more involvement. The involvement is influenced by the ability of an agent to perform a role, as well as the perceived importance of performing such roles from an individual agent's perspective. For instance, some worker agents may adopt additional informal roles in a company (e.g., k for agent $W_2$). Some worker agents may monitor other agents or they may have a charismatic personality and informally establish rules (i.e.~norms) which are executed by the help of other agents (see internalised roles of $ \mathrm{W_1} $). Another example is the case of a knowledge agent who may adopt the informal judging role voluntarily (note the addition of S to $K_2$'s formal current role K). Note this agent could adopt the monitoring role for various reasons (e.g. to help stabilise the rule or to weaken the rule-following by not reporting the violators). %Some Knowledge agents might avoid reporting suspicious behaviours or have less ability to monitor, because they do not have a continuous relationship with employees (small K for $ \mathrm{K_1} $). 

%Also, $ \mathrm{S^K} $ agents might perform some tasks that are not relevant to their formal role. For instance, they may define a local rule and enforce that ($ \mathrm{S^K_1} $) or perform some clerical tasks in their spare time to help the society ($ \mathrm{S^K_2} $). 
Another example is the commanders who may also take additional roles such as K and S (e.g. $C_1$). They may take some informal roles to influence rule change. For instance, even though they may establish a rule, they may feel that they do not have the obligation to follow them and so they may overlook them, hence impacting rule-following for the whole society. These examples described above show how formal and informal roles can shape rule changes in an organisation.%This instance is available in some countries as parliamentary immunity. In an organisation, this can be the case for a sole owner (or main shareholder) of an organisation who tends to overlook the rules (i.e.~does not follow them).

\section{Simulation, algorithms, and parameters}
\label{SimulationAndPArameters}
In this section, first, we discuss the underlying assumptions of this simulation. Then, we provide an overview of two historical societies studied for simulation, namely the English East India Company (EIC) and Armenian merchants of New-Julfa (Julfa). Then we briefly discuss the aspects of these societies that are of interest for us and the simulation procedures used to represent their agents' behaviour in the simulation context.
\subsection{Assumptions}
In societies, the rules that exist may not be honoured by agents. Although, the agents know the existence of such rules, they don't follow them and the agents justify this behaviour through the resolution of cognitive dissonance. \emph{Cognitive dissonance} is defined as tensions formed by conflicts between different cognitions (for instance, one likes to smoke, but loathes to get cancer) \cite{aronson1999social}. These tensions lead to creating some justification for taking one action (quit smoking or continuing). This idea was used to attribute workers' productivity to cognitive dissonance regarding fairness of institutions \cite{Adams1962}. In particular, studies showed that procedural justice (having fair dispute resolution mechanisms) increases public law obedience and cooperation with the police \cite{Sunshine2003}. Also, underpaid or overpaid persons alter their efforts put forth on the system (e.g.~efforts or voluntarily performed tasks) to make the system fairer for themselves \cite{adams1965inequity}. In this work, we consider that agents justify the need for rule change (or don't follow rules),
because they need to resolve this cognitive dissonance (i.e.~they justify not following rules, or the reason to keep following the rules).  %Finally, some experiments indicate that the impact of the fairness on agents' behaviour varies based on their personalities \cite{BuboltzJr2003,Schmitt2005,Schmitt2010}.%  \citet{},\\ \citet{BuboltzJr2003}, and \\ \citet{Schmitt2005,Schmitt2010
\subsection{Societies}
As stated earlier, in this paper, we investigate two long-distance trading societies, namely Armenian merchants of New-Julfa (Julfa) and the East India Company (EIC). The two societies were contemporaneous and shared the same areas for trading products (e.g.~the EIC managers granted Julfans permissions for using the EIC infrastructures \cite{aslanian2007indian}). Also, both societies faced principal-agent problem \cite{Mitnick2011} --- the dilemma where the self-interested decisions of a party (agent) impact the benefits of the other person on whose behalf these decisions are made (principal). 

\textbf{Armenian merchants of New-Julfa ({Julfa}):} Armenian merchants of New-Julfa were originally from old Julfa in Armenia. They re-established a trader society in New-Julfa (near Isfahan, Iran) after their forced displacement in the early 17{th} century \cite{Herzig1991,aslanian2007indian}. %Due to their complicated inheritance rules that created tight bonds within extended families \citep{Herzig1991}, they formed a closed society that was run by strong social norms instead of formal rules. 
   They used commenda contracts (profit-sharing contracts) in the society and also used courts to resolve disputes \cite{Herzig1991,aslanian2007indian}. %The mercantile agents were responsible for buying and selling items and moved among different nodes of the trading network. %The society also had apprenticeship programmes to improve their recruits' skill levels. 
   
%    Based on the historical data, these traders are known to have had a ``merchant school'' \citep[p.171]{aslanian2007indian} around the 1680s. A more general apprenticeship system was active in Julfa and Persia (old Iran) to transfer skills in the society, wherein skilled agents employed the labour of apprentices and trained them instead. In Julfa, apprenticeship took place informally by family members or relatives who hired or recommended trained apprentices. 
   
   \textbf{The English East India Company (EIC):} During the same time, the EIC (AD 1600s-1850s) had a totally different perspective on managing the society. The EIC faced a high mortality rate due to  environmental factors in India. EIC paid fixed wages and fired agents based on their own beliefs about their trading behaviour. Furthermore, EIC's trading period covers the English Civil War (1642--1651), which led to inclusion of some of the senior mangers on the board of directors and granting permission for private trade to the employees (i.e.~trading activities for individuals' self-interests). %We did not consider impacts such as militarisation of the EIC after the 1750s, which was a shift from being traders to being governors, because we are only interested in studying trading aspects of these societies.
    %The incentives of trainers for good training is questionable for reasons like spent time for a task without any provisioned profit (i.e.~like contractors or artisans they trained for immediate income). In pre-modern Britain, guilds were in effect that limited skilled agents from joining the workforce society and made 
    
    In both of these societies agents' meta-roles changed over time. More precisely in EIC, a mercantile or trader agent (W) after gaining experience was promoted to a managerial position to monitor other mercantile agents (K). Also, in EIC, after the English Civil War, managers had the opportunity to be part of the board of directors (C). In Julfa, the promotions took place based on the ageing of the family members (i.e.~agents got promoted from one meta-role to the other gradually). Additionally, in Julfa mercantile agents (W) and heads of families (C) formed the courts ($S$). In this model, we use the EIC dynamics in organisational meta-roles (i.e.~promotion of agents) to make the two systems comparable. Note that this change in dynamics decreases the opportunities for Julfans to revise their rules. However, we know that the rules were deeply honoured by Julfans \cite{aslanian2007indian}. %Below, we discuss two important considerations in these two societies.
    
       \textbf{Environment:} These societies had different mortality rates. On average an EIC agent died before the age of 35 due to harsh environmental circumstances \cite{hejeebu2000microeconomic}. Julfan traders did not face such a situation \cite{aslanian2007indian} and the closed trading society of Julfa would have collapsed under a high mortality rate \cite{Sedigh2019}. %Also, the cost of this mortality rate for companies concerns some fixed costs regarding hiring and sending new agents to the settlements. In addition, companies face costs of lost opportunity, because a dead agent cannot finish the trade.
       
          \textbf{Fairness:} Another difference between the two historical long-distance trading societies is associated with their payment schemes for employees and the adjudication processes (i.e.~use of courts for resolving disputes about suspicious behaviour). EIC rarely employed an adjudication process (e.g.~agents were fired based on their performance because of suspected cheating), and the agents were paid low wages \cite{hejeebu2005contract}. However, in Julfa a mercantile agent was paid based on his performance \cite{aslanian2007indian}. Julfans had adjudication processes to resolve disputes, which considered available evidence \cite{aslanian2007indian}. Though Julfa appears to be fairer than EIC in terms of payment, total fairness can be questioned --- for instance, in the Julfa society, the family wealth and trade was managed and controlled by the eldest brother \cite{Herzig1991}. This rule deprived younger ones from managing their own share of capital.
          
         % \todo{Added this to add a bit of context regarding private trade.} In EIC because the payments were low, some agents traded goods privately as they deemed the system to be unfair and this process boosted their income. This was originally forbidden in the EIC.
          % Such a rule can lead to a partial dissatisfaction regarding the head of a family's decisions for other members.
          
          	\subsection{Algorithms}
	\label{algochap7}
    %\todo{This para doesn't talk about K and C agents - where are these involved in the algorithms} 
    In this subsection, we discuss the procedures employed to simulate role changes within the two societies. The simulation model is split into four distinctive procedures. The first procedure models the societal level of simulation, including creating an initial population and staffing (hiring new mercantile agents) to create a stable population. The second level describes procedures for \textit{mercantile agents}' (W) decision-making and learning the system's parameters. The third level covers the decision-making and learning procedure associated with managers (K). The last procedure is the meta-algorithm that sequentially executes the aforementioned algorithms and updates appropriate parameters. In this algorithm, agent meta-roles may change and the opportunity for institutional dynamics is provided (i.e.~promotion of K agents to C and  changes in institutions).

		\setlength{\textfloatsep}{0.1cm}
		\begin{algorithm}[hbt!]
%		\captionsetup{labelfont={sc,bf}, labelsep=newline}
		\caption{Societal level set-up and initialisation}
		\DontPrintSemicolon
		\label{Initiatealgo}
			\tcc{Intialise the system starting with $ iteration \gets 0 $.}
	%\tcc{Initialise the system}
	Create 500 new agents with $ status \gets new $, random personality aspects, and random parameters\;
	Assign appropriate roles (i.e.~mercantile, managers, and directors) to created agents\; 
				\tcc{$	n $ = deceased and fired agents (mercantile agents and managers) in the previous iteration.}

			The most experienced mercantile agents get promoted to a managerial role\;
		Create $ n $  new agents with: $ status \gets new $, $ Experiene \gets 0 $, and randomly initialise parameters \;
					\tcc{Perceived environment and fairness for inexperienced agents.}
			$PEnvironment \gets RandomUniform (0,1)$\;
			$Fair \gets Random Uniform (0,1)$\;
		\end{algorithm}
	\setlength{\floatsep}{0.1cm}
	
\textbf{Algorithm \ref{Initiatealgo}} shows how the societal level of the system is simulated. In iteration 0, the system is initialised by creating 500 new agents with random parameters (line 1). The roles are assigned to created agents (about 2\% directors, 5\% managers, and the rest mercantile agents).\footnote{These numbers are inspired from the numbers in the EIC \cite{hejeebu2000microeconomic}.} The organisation hires and promotes agents to sustain the number of agents per role --- i.e.~replaces deceased agents (lines 3-4). %As discussed earlier, deceased agents incur some costs for the systems. We take into account such costs by calculating $ Hiring Costs $ (line 3). Furthermore, systems that benefit from apprenticeship programmes face the cost of training in terms of lost opportunity (discussed in Section \ref{subs:param}) per hired agents (lines 4-8). Finally, the rate of return (ROR) for the iteration will be calculated. In calculating the ROR, we assume an agent contributes to the company's ROR based on his experience (line 9).
The rest of the algorithm is executed only for inexperienced agents (i.e.~new recruits). An agent has a completely random understanding of the system's characteristics (lines 5-6). %Besides, untrained agents' rule-understandings and their internal beliefs about the seriousness of violations are modelled using different variances (lines 11-12). The social norms are not learnt during training, because trainees are not in contact with other traders. In other words, the variance of norm understanding for both trained and untrained agents is the same (line 14). Finally, the agent updates his status to experienced so that the system can identify him (line 15).

	\begin{algorithm}[hbt!]
%	\captionsetup{labelfont={sc,bf}, labelsep=newline}
	\caption{Mercantile agent's algorithm (for meta-role W)}
	\DontPrintSemicolon
	\label{factoralgo}
	\SetKwBlock{Begin}{}{}
	\tcc{Update parameters for new recruits.}
	\lIf{Status = New }
	{Set agent's parameter using Algorithm. \ref{Initiatealgo}}
\If{$ Experience > 3 $}{
	%\Repeat
\tcc{Update role and the decision to perform private trade.}
	\If{$  Dissonance(Fair) < DissonThresh $} 
	{\tcc{Agent stops monitoring violations.}
		 Remove $K$ from voluntarily performed roles \;
	
\If{$\Big(\big( Fair < thresh\big) $ \textbf{or} $\big(\dfrac{No.\ PrivateTraderFriends} {   No.\ Friends} < JustifThresh\big)\Big)$}{
		\tcc{Agent decides to perform private trade.}
		$ PrivateTrade \gets OK$}
	}
\tcc{Agent voluntarily collaborates in monitoring.}
\lIf{$  Dissonance(Fair) > DissonThresh $}{
	 Voluntarily perform $K$
}
}
	\tcc{learning;}
	\If{$ Experience>3 $}{
		\tcc{Reporting observed violations;}
		\If{Voluntarily performing K}{
			\tcc{The agent reports some of the cheaters observed.}
			Agent reports connections who impose more costs on the organisation than his tolerance (internalised S).
		}
	}
	
%	\tcc{Agent may die based on the Equation \ref{mortalityenv}.}
Learn parameters and adjust the beliefs about rules\;
	$ Experience \gets Experience + 1 $\;
	\lIf{$Rand(1) \leq MortalityProbability(Experience + 15) $}{Die}
\end{algorithm}	\setlength{\floatsep}{0.1cm}
	
    \textbf{Algorithm \ref{factoralgo}} shows the procedure associated with mercantile agents' decision-making process. Note that in this algorithm $ \#Rnd(x) $ indicates a random number generated in the interval $ (0,x) $. As stated earlier, if the status of the mercantile agent is new, he goes through an initialisation (see Algorithm \ref{Initiatealgo}, lines 3-4). Furthermore, experienced mercantile agents decide on their participation in monitoring by considering cognitive dissonance incurred (based on their perception of institutional fairness and dissonance toleration). They also decide on performing private trade with respect to the perceived fairness and their friends who perform such trades (lines 3-7). %Lines 44-49 model the consequences of permitting private trade. As stated earlier, once private trade is permitted, a drop in trade is evident for three consecutive years (see Section \ref{EICeco}). In our model, each mercantile agent has a threshold by which he decides to leave the organisation (line 44). We also know that after some years, mercantile agents in the EIC faced financial troubles. In other words, after 100 iterations (years after the company's establishment), mercantile agents were faced with a decrease in wages and a drop in private trade's revenue such that they were desperate to pay for their living costs. As a consequence, costs associated with violating the internal beliefs decrease (lines 45-49).
 If the mercantile agent has enough experience and has already decided to collaborate in monitoring, he helps the system to identify violators, based on his interpretation of a fair action (lines 8-9). Finally, the mercantile agent updates his perception of system parameters (e.g.~fairness of the society), increases his experience, and may die (lines 10-12).

%	initialised (this algorithm is called for \textit{New} agents i.e.~the ones who newly joined the system). If system has an apprenticeship programme the hired agents have a better skill and more precise understanding of the system characteristics (lines 1-3). Furthermore, trained agents' has a different belief and rule understanding than their untrained counterparts (line 4). Otherwise, the agent has lower skills and a random understanding of system characteristics (lines 5-7). In addition, agents' understanding of rules and their beliefs towards the rule has a different deviation (lines 8-9). Finally, the agent update his status to experienced so that the algorithm can identify him.
	
	\begin{algorithm}[hbt!]
%	\captionsetup{labelfont={sc,bf}, labelsep=newline}
	\caption{Manager's algorithm (for meta-role K)}
	\DontPrintSemicolon
	\label{MidManager}
	%\SetKwBlock{Begin}{}{}
	%\tcc{AP and Waited will be used in the Apprentice's algorithm.}
	\tcc{Manager reports (and eventually punishes) a number of employees who violate the rules of the organisation beyond its tolerance level. We call the threshold $ TolPunish $.}
	$ PotPunish\gets$ employees with violations more than $ TolPunish $\;
\eIf{The number of members of $ PotPunish > MaxPunish$}{
	\tcc{The manager has a limit for the number of agents he can punish called $ MaxPunish $.}
Punish $ MaxPunish $ out of $ PotPunish $ that have the most violation
}
{
	Punish all $ PotPunish $ members.
} 	
%\tcc{Agent may die based on the Equation \ref{mortalityenv}.}
$ Experience \gets Experience + 1 $\;
\lIf{$Rand(1) \leq MortalityProbability(Experience + 15) $}{Die}
\end{algorithm}	\setlength{\floatsep}{0.1cm}

    \textbf{Algorithm \ref{MidManager}} shows the procedures associated with managers (i.e.~monitoring agents (K)). A manager creates a set that consists of reported violators with unacceptable violations (i.e.~he tolerates violations to some extent, see line 1). Note that the manager reports about the violators and punishes a certain number. If the number of violators exceeds a certain threshold, he punishes the worst violators (lines 2-3). Otherwise, all the violators are punished (lines 4-5). Finally, the agent's experience and age increase, and the agent may die (lines 6-7).
    
	\setlength{\textfloatsep}{0.1cm}
	\begin{algorithm}[hbt!]
%	\captionsetup{labelfont={sc,bf}, labelsep=newline}
	\caption{Meta algorithm}
	\DontPrintSemicolon
	\label{meta}
	%\tcc{AP and Waited will be used in the Apprentice's algorithm.}
	\tcc{Intialise the system starting with $ iteration \gets 0 $.}
	%\tcc{Initialise the system}
	%\tcc{Define a symmetric organisationsl fuzzy number for \textit{{violation labels}}, assign random system parameters to environment based on the environmental characteristics.}
	Create 500 new agents with $ status \gets new $ and random parameters with appropriate roles\;
	%Assign appropriate roles (i.e.~mercantile, managers, and directors) to created agents\;
	%Assign random experience to created agents\;
%	Find $ MaxFriends \times Extraverted  $ number of friends\;
	\tcc{Call algorithms in an appropriate sequence.}
 \Repeat{$ iteration=250 $}{
		Run Algorithm \ref{Initiatealgo}\;
	Run Algorithm \ref{factoralgo}\;
	Run Algorithm \ref{MidManager}\;
	\If{$ iteration = 70 $}{
		Update board of directors (C) with new managers\;
		\lIf {majority support private trade}
		{legalise private trade and reduce wages
		}
		}
	$ iteration \gets iteration + 1  $
}
\end{algorithm}	
	\setlength{\floatsep}{0.1cm}
		
    \textbf{Algorithm \ref{meta}} is the main algorithm that calls the other procedures. In iteration 0, the system is initialised by creating 500 new agents with random parameters. The roles are assigned to created agents (2\% directors (C), 5\% managers (K), and the rest are mercantile agents (W)), and they have 0 years of experience (line 1). Then, 250 iterations corresponding to 250 years, containing specific steps (lines 3-9) are performed (250). The first step is to run the  societal algorithm (i.e.~Algorithm \ref{Initiatealgo}, line 3). Then the algorithm associated with the mercantile agents is run (i.e.~Algorithm \ref{factoralgo}). Finally, the manager's decisions are made using Algorithm \ref{MidManager} (line 5). When the simulation reaches the year that some of the managers in the EIC (who started as mercantile agents) are promoted to the board of directors (i.e.~consequences of the English Civil War, iteration 70), a decision about permitting (or legalising) private trade is made (lines 6-8).
	
\subsection{Parameters}
In this subsection, we discuss the important parameters employed in the simulation (see Table \ref{paramet3}), along with the reasons for choosing specific values for them. Note that we used 250 iterations to reflect the longevity of EIC (it was active with some interruptions and changes in power from 1600 to 1850).  In Table \ref{paramet3}, column `Name' indicates the names of parameters, column `Comment' shows additional information if required, column `Distribution' indicates the probability distribution used for these parameters, and column `Values' indicates the values of parameters estimated for the two societies. Note that these parameters can be modified to reflect other societies.% of the probability distribution or constants.

\begin{table}[htb!]
	\caption{Parameters associated with the model}
\label{paramet3}
				\footnotesize
	\begin{tabular}{|l|l|l|l|}
%\multicolumn{4}{c}%
%{{\bfseries \tablename\ \thetable{} -- continued from previous page}} \\
		\hline	\multicolumn{1}{|c|}{Variable name} & \multicolumn{1}{c|}{Comment} & \multicolumn{1}{c|}{Distribution} & \multicolumn{1}{c|}{Values} \\ \hline% \\ \hline 
		\multicolumn{1}{|l|}{Fairness} & \multicolumn{1}{l|}{Unfair : Fair} & \multicolumn{1}{l|}{Constant} & \multicolumn{1}{l|}{$-0.4:0.6$} \\ \hline
		\multicolumn{1}{|l|}{\begin{tabular}[c]{@{}l@{}}Perception of environment \\ and fairness of system\end{tabular}} & &  Uniform & $(-1,1)$ \\ \hline
	
		\multicolumn{1}{|l|}{Thresholds} & \multicolumn{1}{l|}{\begin{tabular}[c]{@{}l@{}}Dissonance\\ Environment\\ Fired agents\end{tabular}} & \multicolumn{1}{l|}{Uniform} & \multicolumn{1}{l|}{\begin{tabular}[c]{@{}l@{}}$(0,1)$\\ $(0,1)$\\ $(0, 0.3)$\end{tabular}} \\ \hline
	Monitoring& Boolean& Bernoulli &0.5\\ \hline
		\multicolumn{1}{|l|}{\begin{tabular}[c]{@{}l@{}}Permission for private \\ trade\end{tabular}} & \multicolumn{1}{l|}{\begin{tabular}[c]{@{}l@{}}Percent of joined managers\\ who agreed to change\end{tabular}} & \multicolumn{1}{l|}{Constant} & \multicolumn{1}{l|}{$ 70\% $} \\ \hline
		\end{tabular}
\end{table}

\textbf{Fairness:} Note that as discussed earlier, Julfa had fairer institutions than the EIC. We set system fairness values to 0.6 and -0.4 for fair and unfair societies respectively. We believe that neither of these two societies were totally fair or unfair (e.g.~EIC managers justified the firing of agents that indicates there has some effort towards fairness).
% However, some attempts in the EIC to convince the agents and inheritance rules in the Julfa convinces us to consider that the EIC score for being unfair is 0.6 (i.e.~it is -0.4 fair). Also the score for fairness of Julfa is 0.6.

%	0.001 + 0.00015 ^ (Env+? + 0.01)
%	0.067 - (0.064 / (1 + ((Env+? / 0.55) ^ 3.38)))

\textbf{Perceived characteristics:} Because of lack of prior experience, the new agents have a totally random understanding of social characteristics.

\textbf{Thresholds:} These are the numbers that reflect an agent's tolerance of different aspects and characteristics of the system. All these thresholds are generated at random except for firing. For the proportion of fired agents, we assume that a manager would fire 30\% of the suspected employees. %This 30\% mirrors the observation that initially around 30\% of agents cheated\footnote{In both societies agents' dismissal was not very common \cite{aslanian2007indian,hejeebu2005contract}.}.% (these numbers are presented in the next section). 

\textbf{Monitoring:} In the model, a recruit may voluntarily decide to participate in monitoring --- we use a random boolean generator to represent this.% However, as discussed earlier, these agents may change their decisions based on the situation.

\textbf{Permission for private trade:} In this simulation, we assume that permission is granted if more than 70\% in the board of directors agree to such a decision (i.e.~8 out of 11). %We used this relatively high proportion to model the impact of their negotiation power with other directors (i.e.~a strong attitude towards such a practice should be introduced to the board of directors). 

Furthermore, we parametrise the agents' learning as follows. Agents discount information using a weight of 30\% for the past. This reflects the importance of recent information for agents.

\section{Results}
\label{results}
In this section, we describe the simulation results considering \textit{four} different combinations of two characteristics, namely a) environmental circumstances and b) fairness of institutions. With two different values for each of these characteristics, four combinations are possible (see Table \ref{specdyn}). 

\begin{table}[hbt!]
	\centering
	%\footnotesize
	\caption{System specification based on different characteristics}
	\label{specdyn}
	\begin{tabular}{lcccc}
		\hline
		Characteristics & {\begin{tabular}[c]{@{}c@{}}$E_0F_0$\\ (EIC)\end{tabular}} & {$E_0F_1 $} & {$E_1F_0 $} &
		   
		  			{\begin{tabular}[c]{@{}c@{}}$E_1F_1 $\\ (Julfa)\end{tabular}} \\\hline
		Environment	 & \ding{55} & \ding{55} & \ding{51} & \ding{51} \\
	Fairness		 & \ding{55} & \ding{51} &  \ding{55} & \ding{51} \\\hline
		
	%	\multicolumn{17}{l}{{$^2$}A representative for EIC} \\
	%	\multicolumn{17}{l}{{$^3$}A representative for Julfa}
	\end{tabular}
\end{table}

%Table \ref{specdyn} indicates the characteristics for the 4 simulated societies and societies they represent. 

The configurations (i.e.~societies) are identified by the first letter of the characteristics, namely \textit{E} and \textit{F} that are representatives of the \textbf{e}nvironmental characteristic of (E) and \textbf{f}airness of the institutions (F), respectively. A tick indicates that the society possesses such an attribute, and a cross indicates the society does not possess such an attribute. In this table, we gradually change characteristics of the EIC ($E_0F_0$) to get closer to Julfa ($E_1F_1 $), to examine their effects on the success of these societies. We utilised \textit{NetLogo} to perform our simulations \cite{netlogo}. We also used 30 different runs for each set-up and then averaged their results. Finally, note that the patterns observed in simulation results are compared to the patterns reported from the EIC and Jufla, because we had access to the qualitative data.

\subsection{Permissions for private trade}
Table \ref{private} presents the percentage of simulation runs (out of 30) where the permission for private trade was granted (see row ``Permission granted''). Note that this change in rule (granting of permission) happened due to changes in agents' meta-roles where a mercantile agent progresses to the board of directors (and advocates the decision to permit private trade). As can be seen from the results, both unfair societies ($F_O$) had higher percentage of runs where the private trade is permitted ($>50\%$), although with a large difference (93\% and 57\% respectively). In fair societies, none of the runs resulted in private trade being approved. This result mirrors the evidence from Julfa. In Julfa, mercantile agents (W) and peripheral managers (K) were the ones who eventually ran the family business (C). Also, mercantile agents and managers made decisions regarding violations and acted as juries in certain courts \cite{aslanian2007indian}. The aforementioned situation, combined with keeping private trade illegal \cite{Herzig1991}, indicate that this rule was socially accepted.\footnote{Because in none of the simulation runs of Julfa the permission was granted, we believe that using a similar dynamics to Julfa would not change the results} Also, we know that in the EIC, the permission for private trade was granted once the managers had the opportunity to be part of the board of directors\cite{erikson2014between}.  

\begin{table}[hbt!]
	\centering
	%\footnotesize
	\caption[Percentage of runs where private trade was permitted]{Percentage of runs where private trade was permitted (out of 30 runs).}
	\label{private}
	\begin{tabular}{|c|c|}
		\hline
	\textbf{Societies} & \textbf{Permission granted for private trade} \\ \hline
		$E_0F_0$ (EIC)& 93\%   \\ \hline
		$E_0F_1$& 0\%   \\ \hline
	$E_1F_0$ & 57\%   \\ \hline
		$E_1F_1$ (Julfa)& 0\%   \\ \hline
	\end{tabular}
\end{table}

%However, granting permission for private trade in unfair societies is different. Keeping in mind that this permission is granted if more than 70\% managers agree (see permissions for private trade in Section \ref{subs:param}), additional characteristics are also important. For instance, societies with a favourable environment and apprenticeship programmes are less likely to grant such permissions. Overall, fair institutions and having apprenticeship programmes deters agents from performing private trade. 

	\begin{figure}[hbt!]
		\centering
		%	trim=left bottom right top
	%	\rotatebox{270}{
			\includegraphics[trim=0cm 2.3cm 0cm 0cm,clip,
			width=.95\textwidth]{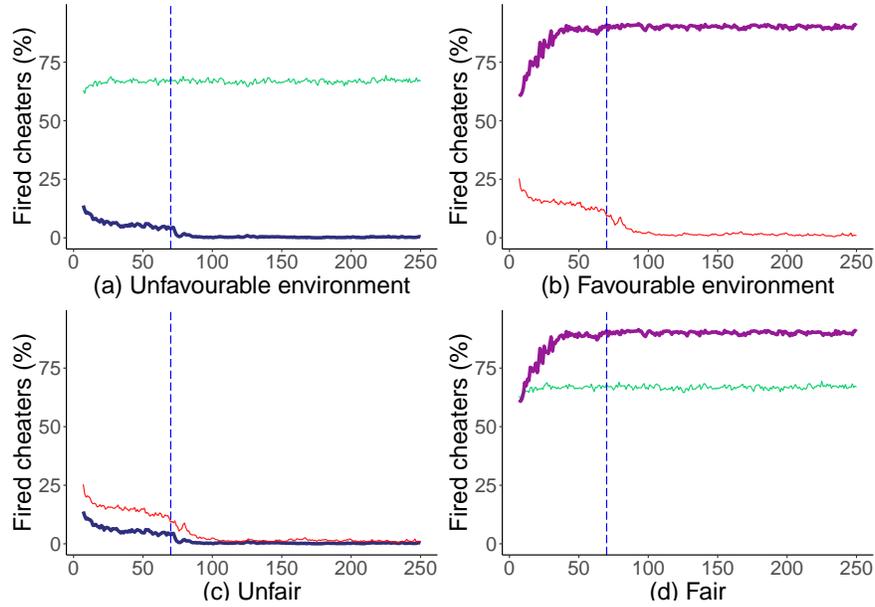}
	%	}
		\caption{Monitoring strength and firing in simulated societies.}
		\label{monitor}
	\end{figure}%\cmt{Colour-coding borderline ok. Have a look in b/w and see what you think.}

\subsection{Fired violators (monitoring strength)}

 Now we discuss the impact of aforementioned two characteristics on the monitoring strength of the system (see Figure \ref{monitor}). Figures \ref{monitor}a-\ref{monitor}d present the percentage of the cheating agents fired. In these figures, the y-axis indicates the percentage of fired cheaters. As can be seen, the most fired agents belong to society $ E_1F_1 $ and then $E_0F_1$. These indicate the importance of fairness of institutions on the system's monitoring strength.
 
 This impact that we see in Figure 3 is a consequence of two informal roles performed by agents, namely a) mercantile agents (W) that monitor and report suspicious behaviours (internalised K) to managers (formal K), and b) managers (K) who interpret rules based on the situations and tolerate some behaviours (S). For example, managers who think the system isn't fair, may not report the cheating behaviour of agents (agents who are involved in private trades). And these same managers who become a part of the board of directors allow for these private trades to happen legally (but with the reduction in wages further, though). %However, finding a pattern that applies to all societies with shared characteristics is difficult.
%Before analysing the results, one should keep in mind that the fired proportion alone presents the societal reaction to violations 
Also, in organisations with unfair institutions, after granting permissions for private trade (year 70), agents' collaboration in monitoring the cheaters (for theft etc.) decreases. Note that the evidence for interpretation of the rules can be found in EIC managers' correspondence\footnote{For instance, in the early years, some managers defended mercantile agents' private trade by stating:
	``if some tolleration [sic] for private trade be not permitted none but desperate men will sail our ships'' %(Factory Records: Miscellaneous, I, 26, 18, February 1620, p.~87 \cite{chaudhuri1965english}) 
	\cite{chaudhuri1965english}.}. Also, these results mirror the evidence of rare cheating and successful monitoring mechanisms in Julfa \cite{aslanian2007indian} and the popularity of cheating and collusion in the EIC \cite{chaudhuri1965english}.
%In connection with the statistical test, note that we are not interested in comparing the percentage of agents fired irrespective of the impact of the percentage of cheaters. In other words, we are interested in the percentage of cheaters who are fired, as opposed to the percentage of those fired compared to the percentage of agents. Note that the latter percentages are misleading when we compare societies with a high number of cheaters versus societies with a low number of cheaters. In other words, although most of the historical information concentrates on the percentage fired, comparing agents fired in the two societies with totally different percentages of cheaters seems to be misleading. Because of that, we have used a partial correlation of the Kendall's test to remove the effect of the percentage of cheaters.

\section{Discussion and concluding remarks}
\label{conc}

This study has presented an extension of the BDI cognitive architecture to investigate its interaction with agents meta-roles. Also, using this extension, the study has investigated the impact of a combination of a) dynamics in agents' roles and b) the institutional characteristics (i.e.~mortality rate and fairness) on organisational rule dynamics (i.e.~change of rule). As the role of individuals changes (e.g.~W to K), their beliefs formed based in their previous role impacts their new decisions. Finally, our study has used the evidence from empirical studies to simulate two historical long-distance trading societies, namely Armenian merchants of new Julfa (Julfa) and the English East India Company (EIC) and has demonstrated what may cause rule changes (i.e.~role change and institutional characteristics).

%Also, to simulate the historical cases, we considered two characteristics of these systems, namely environmental characteristics and fairness of institutions. 
The simulation results mirrored historical evidence. It has shown that the fairness of institutions is a pivotal characteristic to drive their stability (i.e.~avoiding revisions in rules) and in facilitating agents' collaboration in monitoring each other's behaviours. These results (i.e.~changes in rules and weak monitoring and reporting) mirror concerns in the modern context about the division of ``rules into the two categories of rules-in-use and rules-in-form'' \cite{rulesinuse}. For instance, it is noted that rules-in-use (followed rules) in some provinces of Canada might have been rules-in-forms (unfollowed rules that do not have any effect on behaviour) in others \cite{rulesinuse}. There exist some obstacles in a law in becoming a rule-in-use \cite{DANIEL2017}. An instance of this obstacle is the activities of monitoring agents who interpret the law differently and thus hamper its effectiveness (e.g. through not monitoring violations) and hence can aid the formation of new rules similar to what has been observed in results from Table \ref{private} and Figure \ref{monitor}.

A future extension of the study, will involve detailed examination of the interaction between other modules of the cognitive architecture presented in Figure \ref{architec}. Also, the simulation can be extended to take account of other characteristics of these historical societies, such as the personalities of agents, to provide a more fine-grained model.% improve our understanding of 
%\noindent 

% the environments 'definition', 'lemma', 'proposition', 'corollary',
% 'remark', and 'example' are defined in the LLNCS documentclass as well.
%

 \bibliographystyle{splncs04}
\bibliography{ref}

\end{document}